\documentclass{article}

\PassOptionsToPackage{numbers}{natbib} 

\usepackage[preprint]{neurips_2025}

\usepackage{amsmath,amsfonts,bm}

\def\eqref#1{equation~\ref{#1}}

\def\1{\bm{1}}

\DeclareMathAlphabet{\mathsfit}{\encodingdefault}{\sfdefault}{m}{sl}
\SetMathAlphabet{\mathsfit}{bold}{\encodingdefault}{\sfdefault}{bx}{n}

\usepackage{hyperref}
\usepackage{url}
\usepackage{graphicx}
\usepackage{pifont} 
\usepackage{algorithm}
\usepackage{algpseudocode}
\usepackage{color}
\usepackage{xcolor}
\usepackage{booktabs}
\newcommand{\greenuline}[1]{\textcolor{green}{\underline{\textcolor{black}{#1}}}}
\usepackage{listings}
\usepackage{multirow}
\usepackage{makecell}
\usepackage{tikz}
\usepackage{caption}
\usepackage{cancel}
\usepackage{wrapfig}  

\title{
Unsupervised Radar Point Cloud Enhancement via Arbitrary LiDAR Guided Diffusion Prior \\
}

\author{%
  Yanlong Yang \\
  James Cook University\\
  \texttt{yanlong.yang@my.jcu.edu.au} \\
  \And
  Jianan Liu \\
  Momoni AI \\
  \texttt{jianan.liu@momoniai.org} \\
  \AND
  Guanxiong Luo \\
  University Medical Center Göttingen \\
  \texttt{guanxiong.luo@med.uni-goettingen.de} \\
  \And
  Hao Li \\
  Heidelberg University \\
  \texttt{hao.li@med.uni-heidelberg.de} \\
  \And
  Euijoon Ahn \\
  James Cook University \\
  \texttt{euijoon.ahn@jcu.edu.au} \\
  \And
  Mostafa Rahimi Azghadi\\
  James Cook University \\
  \texttt{mostafa.rahimiazghadi@jcu.edu.au} \\
  \And
  Tao Huang \\
  James Cook University \\
  \texttt{tao.huang1@jcu.edu.au} \\
}

\begin{document}

\maketitle

\begin{abstract}
In industrial automation, radar is a critical sensor in machine perception. 
However, the angular resolution of radar is inherently limited by the Rayleigh criterion, which depends on both the radar’s operating wavelength and the effective aperture of its antenna array.
To overcome these hardware-imposed limitations, recent neural network-based methods have leveraged high-resolution LiDAR data, paired with radar measurements, during training to enhance radar point cloud resolution.
While effective, these approaches require extensive paired datasets, which are costly to acquire and prone to calibration error.
These challenges motivate the need for methods that can improve radar resolution without relying on paired high-resolution ground-truth data.
Here, we introduce an unsupervised radar points enhancement algorithm that employs an arbitrary LiDAR-guided diffusion model as a prior without the need for paired training data. 
Specifically, our approach formulates radar angle estimation recovery as an inverse problem and incorporates prior knowledge through a diffusion model with arbitrary LiDAR domain knowledge.
Experimental results demonstrate that our method attains high fidelity and low noise performance compared to traditional regularization techniques. Additionally, compared to paired training methods, it not only achieves comparable performance but also offers improved generalization capability.
To our knowledge, this is the first approach that enhances radar points output by integrating prior knowledge via a diffusion model rather than relying on paired training data.
Our code is available at \url{https://github.com/yyxr75/RadarINV}.
\end{abstract}

\section{Introduction}

Radio Detection and Ranging (Radar) technology has been extensively used in robot and related technology.
In traffic monitoring, radar systems have been employed for vehicle detection \citep{palffy2020cnn} and tracking \citep{liu2023smurf}. In the realm of intelligent driving, radar has been integrated with other sensors like LiDAR \citep{yang2022ralibev} and cameras \citep{zheng2023rcfusion} to enhance object detection \citep{xiong2022contrastive} and cooperative perception \citep{huang2023v2x}. 
However, limited by the number of antennas and hardware noise, the angular resolution of radar is restricted.

To improve radar angular resolution, Multiple Input Multiple Output (MIMO) techniques can increase the number of antennas \citep{bliss2003multiple}, but enhancing resolution with fixed hardware remains challenging. Deep learning methods, such as supervised learning, map sparse radar data to dense Light Detection and Ranging (LiDAR) data using neural networks \citep{jin20232d, kim2024pillargen}, relying on paired data for training to enhance resolution. In addition, due to the sparsity of radar inputs, these methods show limited performance, prompting the use of richer radar signals as input \citep{prabhakara2023high, cheng2021new}. However, these methods yield limited performance, poor generalizability and low robustness.

To address this, methods based on generative models such as diffusion models \citep{song2020score, ho2020denoising} 
have been proposed. These methods usually set LiDAR point cloud as target data domain and radar point cloud \citep{wu2024diffradar, luan2024diffusion, wu2025diffusionbasedmmwaveradarpoint, zheng2025r2ldmefficient4dradar} or radar range-azimuth heatmap \citep{zhang2024towards}, which is a 2D radar signal map encoding range and azimuth as intensity values, as conditioning inputs. 
During the training of generative models, paired training is also conducted \citep{rombach2022high}, requiring the model to output corresponding LiDAR data when given the radar as a condition.
By learning the distribution and structure of the target data, diffusion models based on input radar point clouds can produce high-quality, diverse new samples, often outperforming direct CNN model mappings in many tasks.
Despite this, the aforementioned supervised methods still require one-to-one correspondence between radar and LiDAR data, which is impractical in scenarios where LiDAR sensors are absent or not properly aligned—a common limitation in cost-sensitive vehicles. In practice, many production-level vehicles, especially those targeting the mid- to low-end market, are equipped only with radar and cameras to support basic Advanced Driver Assistance Systems (ADAS) functionalities such as collision warning, adaptive cruise control, and lane-keeping assistance. These systems prioritize affordability and robustness over high-resolution 3D perception, making it infeasible to rely on expensive, power-hungry, and maintenance-intensive LiDAR sensors. Furthermore, even in vehicles equipped with LiDAR, environmental factors like dirt, rain, or snow can cause data dropout or misalignment, further complicating the use of supervised methods that depend on precise radar-LiDAR pairing.


To address this challenge, we propose an unsupervised radar point cloud enhancement method guided by an arbitrary LiDAR-based diffusion prior. For this, we formulate radar angle estimation recovery as an inverse problem by explicitly modeling the radar imaging process and solving it through a Bayesian framework.

The primary contributions of this paper are as follows:
\begin{itemize}
    \item We propose a novel unsupervised method for radar point cloud enhancement that leverages a diffusion model guided by arbitrary LiDAR domain knowledge, without requiring paired radar-LiDAR training data. To the best of our knowledge, this is the first successful application of such an approach in the context of radar point enhancement.

    \item Specifically, a latent diffusion model is trained to reconstruct LiDAR data of RADIal dataset. Such diffusion model driven by LiDAR domain knowledge is treated as a prior distribution in the inference procedure. Once such prior is trained, LiDAR data is not required anymore. As a result, the enhanced radar point cloud is generated as a posterior sampling by incorporating radar imaging forward function as likelihood.
    
    \item We evaluate our method on real-world RADIal datasets, showing it achieves comparable performance to supervised approaches, even without requiring paired training data. We further validate our method on cross-dataset testing using K-Radar, with a fixed diffusion prior trained on RADIal, demonstrating superior generalizability over previous approaches.
\end{itemize}

\section{Related Works}

\paragraph{Discriminative Model-based Radar Points Enhancement:}
With advancements in deep learning technology, researchers have leveraged the robust fitting capabilities of neural networks to directly train models that map inputs to outputs, known as discriminative models. Some studies have utilized radar point cloud data, structured in either 2D \citep{jin20232d} or 3D formats, and processed through UNet networks \citep{kim2024pillargen} to train the mapping parameters from radar inputs to LiDAR outputs. However, the effectiveness of these methods is constrained by the sparse information present in radar point clouds. To improve performance, subsequent research has replaced radar point clouds with radar signals as input, such as time-domain signals \citep{jiang20234d}, range-Doppler signals \citep{cheng2021new, cheng2022novel}, range-azimuth signals \citep{prabhakara2023high}, or range-Doppler-elevation-azimuth signals \citep{han2024denserradar, roldan2024see, roldan2024deep}. This approach also employs CNN/Transformer-based network architectures to perform the mapping. Although these radar signals introduce higher noise, they contain richer information, leading to improved enhancement outcomes. Nonetheless, these models still face significant challenges with generalization and lack the ability to generate detailed outputs.

\paragraph{Generative Model-based Radar Points Enhancement:}
To improve model generalization, approaches based on generative models have been introduced. These models are capable of producing data within a specific domain by learning a mapping from Gaussian distributions to the target data’s statistical distributions \citep{song2020score}. The diffusion model, currently the state-of-the-art among generative models, achieves this by progressively adding noise to the original data until it approximates Gaussian noise, and then reversing this process through a Markov chain \citep{ho2020denoising}. During training, neural networks learn the noise parameters at each step of the noise-addition process, and during denoising, they gradually remove the noise to generate new data within the target domain \citep{rombach2022high}. Building on this framework, some studies have designated LiDAR as the target data domain for enhanced point clouds while using radar as the condition \citep{wu2024diffradar}, thereby training various conditional diffusion models. These investigations enhance inference performance by incorporating additional consistency constraints during training \citep{zhang2024towards} or by defining biased mapping domains \citep{luan2024diffusion}. However, these approaches rely on well-paired training data, which greatly limits model flexibility and broad applicability. 
To address this limitation, we turn to an alternative formulation: treating radar point cloud enhancement as an inverse problem \citep{song2023solving, rout2024solving}. These approaches leverage the generative capabilities of diffusion models to reconstruct high-quality outputs from incomplete or degraded inputs, demonstrating versatility across domains such as medical imaging \citep{luo2023bayesian, luo2024autoregressive}, computer vision \citep{chung2023diffusion, NEURIPS2022_a48e5877}, and signal restoration \citep{lemercier2024diffusionmodelsaudiorestoration}. Existing diffusion-based Bayesian methods have primarily addressed single-modal, synthetic datasets. In contrast, our work extends these techniques to enhance radar point clouds, introducing a novel application in real-world, low-fidelity radar scenarios. This effort represents one of the first explorations of diffusion models guided inverse problem in this domain, highlighting their potential for practical, multi-modal sensing challenges.

\section{Proposed Methods}

This section begins by outlining the radar angle measurement model. Next, it details the method for solving the radar angle estimation inverse problem using the Bayesian theorem. Finally, it explains incorporating a diffusion model.

\subsection{Radar Angle Measurement Forward Model}

\begin{wrapfigure}{r}{0.6\textwidth}
    \centering
    \includegraphics[width=\linewidth]{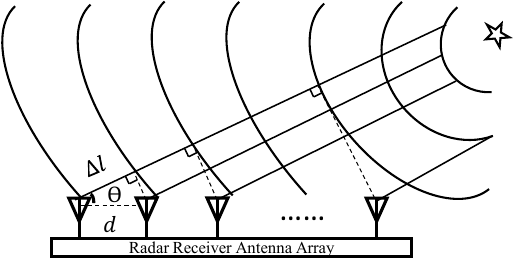}
    \caption{\textbf{Schematic diagram of the radar angle measurement principle}. The five-pointed star symbolizes objects detected in space. As the object-radar distance greatly exceeds the inter-antenna distance, the incident electromagnetic field is treated as a plane wave, arriving at each antenna at the same angle.\vspace{-15pt}}
    \label{fig2: radar measurement fucntion}
\end{wrapfigure}

As depicted in Figure \ref{fig2: radar measurement fucntion}, the electromagnetic waves reflected from an object arrive at the radar's different antennas 
at different times. This time shift results in a phase shift of the waves \citep{friedlander1979inverse, rosen2000synthetic}. Assuming the right antenna is the first antenna. The electromagnetic wave received by the $k$-th antenna is described as:
    $\vec{\textbf{E}}(t)=E_0e^{j\vec{\omega} (t-k\frac{\triangle l}{c})}, $
where $E_0$ and $\vec{\omega}$ represent the amplification and angular frequency of the electromagnetic wave at time $t$. $c$ is the speed of light and $\triangle l$ is the 
path difference, which can be calculated by
    $\triangle l = d\cos(\theta)$,
where $d$ is the constant distance between antennas in the antenna array. Here, $\theta$ is different for different antennas. However, it is often assumed that the incident electromagnetic wave is parallel to the antenna because the antenna spacing distance, $d$ is much smaller than the distance from the object to the radar \citep{logvin2002methods, cheney2009problems}. Therefore, the signal phases received by the antenna array can be expressed as

\begin{equation}
    \boldsymbol{s}_i=\left[0, \frac{d\cos (\theta_i)}{c}, 2 \left( \frac{d\cos(\theta_i)}{c} \right), \cdots, (N-1) \left( \frac{d\cos(\theta_i)}{c} \right) \right]^\text{T}, i \in [1,M].
\end{equation}

Here, $N$ is the number of the antenna. $M$ is the number of objects that separate in space with angles $\theta_i$ to the radar antennas.
The multiple objects signal $\boldsymbol{Y}$ received by the antenna array is
\begin{equation}
\label{radar_measurement_function}
    \boldsymbol{Y} = \left[\boldsymbol{s}_1, \boldsymbol{s}_2, \cdots, \boldsymbol{s}_M\right]+ \boldsymbol{H} = A(\boldsymbol{x})+\boldsymbol{H}.
\end{equation}

Where $\boldsymbol{H}$ is the noise. $\boldsymbol{Y}, A(\boldsymbol{x})=\left[\boldsymbol{s}_1, \boldsymbol{s}_2, \cdots, \boldsymbol{s}_M\right]$ and $\boldsymbol{x}=\left[\theta_1, \theta_2, \cdots, \theta_M\right]$ are generally known as measurement system forward function and unknown state vector in inverse problem theory \citep{chung2023diffusion}.
Based on this, the radar imaging process is reformulated for parallel computing, with further details provided in Appendix \ref{app_c}.  


In real-world scenarios, environmental objects such as roads or building walls are continuous, and each point on such structures can reflect electromagnetic waves. Therefore, the number of potential reflectors $M$ is theoretically infinite. However, the number of antennas $N$ is much smaller than $M$, making Eq.~(\ref{radar_measurement_function}) an underdetermined system of equations. Estimating the unknown state vector $\boldsymbol{x}$ from the measurement vector $\boldsymbol{Y}$ thus constitutes an ill-posed inverse problem.

\subsection{Angle Estimation Reconstruction via Bayesian Estimation}

To solve the inverse problem described above, the Bayesian Estimation is applied \citep{luo2020mri}. Assuming noise $\boldsymbol{H} \sim \mathcal{N}(0, \boldsymbol{I})$ in Eq. (\ref{radar_measurement_function}), then the likelihood function is depicted as
\begin{equation}
    \label{likelihood}
    p(\boldsymbol{Y}|\boldsymbol{x}) = \mathcal{N}(A(\boldsymbol{x}), \boldsymbol{I}).
\end{equation}
In our case, the measurement $\boldsymbol{Y}$ from radar is known, we need to estimate the $\boldsymbol{x}$ with maximum probability from the limited measurement and the likelihood function, that is the posterior probability $p(\boldsymbol{x}|\boldsymbol{Y})$.
According to Bayesian theorem, the posterior probability $p(\boldsymbol{x}|\boldsymbol{Y})$ is proportional to the multiplication of prior probability $p(\boldsymbol{x})$ and likelihood function $p(\boldsymbol{Y}|\boldsymbol{x})$ as below: 
\begin{equation}
    \label{bayesian_theorem}
    p(\boldsymbol{x}|\boldsymbol{Y}) = \frac{p(\boldsymbol{Y}|\boldsymbol{x})p(\boldsymbol{x})}{p(\boldsymbol{Y})} \propto p(\boldsymbol{Y}|\boldsymbol{x})p(\boldsymbol{x}). 
\end{equation}



The likelihood function \( p(\boldsymbol{Y}|\boldsymbol{x}) \) enables direct application of Maximum Likelihood Estimation (MLE) to estimate the optimal distribution of \( \boldsymbol{x} \). However, this can lead to high variance or sub-optimal solutions. Regularization methods like \( L_1 \)/\( L_2 \) \citep{shkvarko2016solving} improve stability but assume \( p(\boldsymbol{x}) \) follows a Gaussian or Laplacian distribution, limiting their effectiveness. Instead, a diffusion model-based prior is introduced for \( p(\boldsymbol{x}) \).
The prior \( p(\boldsymbol{x}) \) is derived from the LiDAR data distribution, where each LiDAR point is a point-object with a relative angle \( \theta \), as in Eq. (\ref{radar_measurement_function}). This spatial prior from LiDAR enhances the density and fidelity of the output radar points.

\begin{figure}[t]
\begin{center}
\includegraphics[width=1\linewidth]{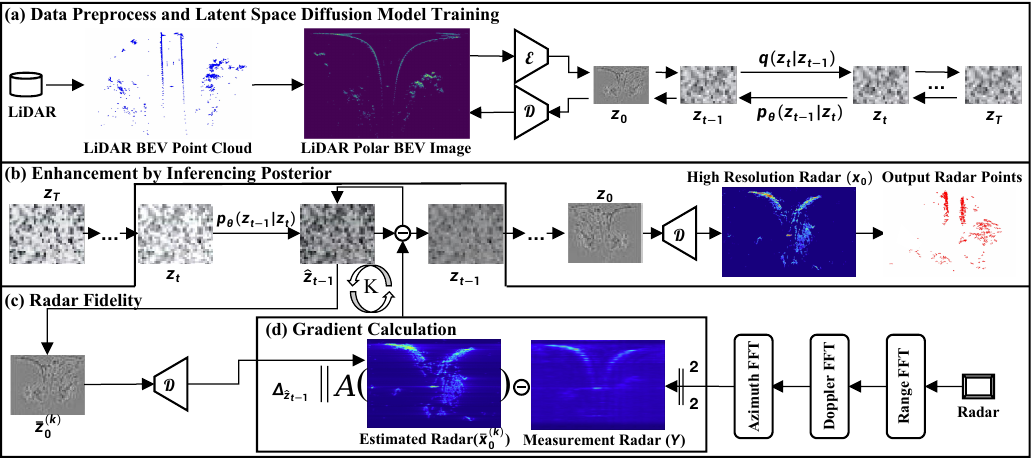}
\caption{
\textbf{Diffusion prior for unsupervised radar point cloud enhancement.}
Let $z_t$ be the latent at step $t$, with $\mathcal{E}$, $\mathcal{D}$ as the autoencoder. The forward and reverse diffusion are modeled by $q(\boldsymbol{z}_t \mid \boldsymbol{z}_{t-1})$ and $p_{\theta}(\boldsymbol{z}_{t-1} \mid \boldsymbol{z}_t)$. ${A}(\cdot)$ is the radar measurement system forward function; \textcircled{-} denotes element-wise subtraction.
(a) A diffusion model is trained on LiDAR latents to learn a prior.
(b) Enhancement samples from the posterior combining LiDAR prior with radar input.
(c) Each reverse step adds an $L_2$ gradient to enforce radar fidelity.
(d) This gradient is computed between ${A}(\bar{\boldsymbol{x}}_0^{(k)})$ and radar data $\boldsymbol{Y}$ w.r.t. $\hat{\boldsymbol{z}}_{t-1}$.
(b) to (d) constitute the inference process, which only requires \( p_{\theta}(\boldsymbol{z}_{t-1} \mid \boldsymbol{z}_t) \) estimated from a trained model in (a), no more training is needed. The details of process (a), (b) are described in section \ref{Diffusion_Forward_Process} and \ref{Diffusion_Reverse_Process}, (c) and (d) are further explained in section \ref{Using_Diffusion_Model_as_a_Prior_for_Enhancement}.}

\vspace{-20pt}
\label{fig2_method_framework}
\end{center}
\end{figure}

\subsection{Apply Diffusion Model as Prior}
We infer the posterior in Eq.~(\ref{bayesian_theorem}) by employing a diffusion model as the prior $p(\boldsymbol{x})$.
The traditional gradient descent algorithm applies the calculated gradient when 
$L_1$ or 
$L_2$ regularization is used, while the diffusion model utilizes a neural network to predict the gradient in each reverse step. Figure \ref{fig2_method_framework} depicts the proposed method framework.

\subsubsection{Diffusion Forward Process}
\label{Diffusion_Forward_Process}
Following the forward diffusion process as depicted in Figure \ref{fig2_method_framework}.(a), we gradually add Gaussian noise to the latent representation of LiDAR data to model its distribution. Given a latent feature \( \boldsymbol{z}_0 = \mathcal{E}(\boldsymbol{x}_0) \) extracted from the LiDAR data \( \boldsymbol{x}_0 \) using encoder \( \mathcal{E} \), noise is added over \( T \) steps. At each step \( t \), the noisy latent \( \boldsymbol{z}_t \) is computed as:
\begin{equation}
    \boldsymbol{z}_t = \sqrt{\bar{\alpha}_t} \boldsymbol{z}_0 + \sqrt{1 - \bar{\alpha}_t} \eta,
\end{equation}
where \( \bar{\alpha}_t \) is a scheduling parameter, and \( \eta \sim \mathcal{N}(0, \mathbf{I}) \). A neural network \( \epsilon_\delta(\boldsymbol{z}_t, t) \), parameterized by \( \delta \), is trained to predict the noise \( \epsilon \), minimizing the loss:
\begin{equation} \label{eq_diffusion_loss}
    \mathcal{L}_\delta = \|\epsilon - \epsilon_\delta(\boldsymbol{z}_t, t)\|^2.
\end{equation}
This process enables the model to learn the LiDAR data distribution in the latent space.

\subsubsection{Diffusion Reverse Process}
\label{Diffusion_Reverse_Process}
The reverse diffusion process aims to denoise the latent representation, reconstructing \( \boldsymbol{z}_0 \) from pure noise \( \boldsymbol{z}_T \sim \mathcal{N}(0, \mathbf{I}) \). At each step \( t \), a trained network \( \epsilon_\delta(\boldsymbol{z}_t, t) \) predicts the noise in \( \boldsymbol{z}_t \). The denoising step follows:
\begin{equation}
    \boldsymbol{z}_{t-1} \approx \frac{1}{\sqrt{\alpha_t}} \left( \boldsymbol{z}_t - \frac{1 - \alpha_t}{\sqrt{1 - \bar{\alpha}_t}} \epsilon_\delta(\boldsymbol{z}_t, t) \right) + \sigma_t \boldsymbol{z},
\end{equation}
where \( \alpha_t \), \( \bar{\alpha}_t \), and \( \sigma_t \) are scheduling parameters, and \( \boldsymbol{z} \sim \mathcal{N}(0, \mathbf{I}) \). This iterative process, shown in Figure \ref{fig2_method_framework}.(b), samples from the posterior \( p_\theta(\boldsymbol{z}_{t-1} | \boldsymbol{z}_t) \), gradually recovering the latent feature \( \boldsymbol{z}_0 \), which is then decoded into enhanced radar data \( \boldsymbol{x}_0^{(\kappa)} = \mathcal{D}(\boldsymbol{z}_0) \).

\subsubsection{Using Diffusion Model as a Prior for Enhancement}
\label{Using_Diffusion_Model_as_a_Prior_for_Enhancement}

\begin{algorithm}
\caption{Gradient Descent for Angle Estimation with Diffusion Model}
\label{alg1}
\begin{algorithmic}[1]
\Require Measurements $\boldsymbol{Y}$, measurement function $A(\cdot)$, total time steps $T$, diffusion coefficient $\lambda_{\text{diff}}$, measurement step size $\zeta$, measurement scale $\gamma$, measurement update iterations $K$
\Ensure Refined angle estimate $\hat{\boldsymbol{x}}$

\State $\boldsymbol{z}_T \sim \mathcal{N}(\boldsymbol{0}, \boldsymbol{I})$ \Comment{Initialize with random noise}
\For{$t = T$ to $1$}
    \State // Calculate diffusion model update
    \State $\boldsymbol{\mu}_\delta \gets S_\theta(\boldsymbol{z}_t, t)$ \Comment{Predict mean using neural network}
    \State $\boldsymbol{\hat{z}}_{t-1} \gets \boldsymbol{z}_t + \lambda_{\text{diff}} \boldsymbol{\Sigma}_\delta(\boldsymbol{z}_t, t)^{-1} (\boldsymbol{\mu}_\delta - \boldsymbol{z}_t)$
    
    \State // Iterative measurement model update
    \State $\boldsymbol{\hat{z}}_{t-1}^{(0)} \gets \boldsymbol{\hat{z}}_{t-1}$
    \For{$k = 1$ to $K$}
        \State $\bar{\boldsymbol{z}}_0^{(k)} \gets \frac{1}{\sqrt{\alpha_t}}(\boldsymbol{\hat{z}}_{t-1}^{(k-1)}-\sqrt{1-\alpha_t}S_\delta(\boldsymbol{\hat{z}}_{t-1}^{(k-1)},t))$ \Comment{Tweedie's formula}
        \State $\boldsymbol{x}_0^{(k)} \gets \mathcal{D}(\bar{\boldsymbol{z}}_0^{(k)})$ \Comment{Decode to image space}
        \State $\nabla_{\text{meas}}^{(k)} \gets \nabla_{\boldsymbol{z}} \| \gamma\boldsymbol{Y}-{A}(\boldsymbol{x}_0^{(k)}) \|_2^2$
        \State $\boldsymbol{\hat{z}}_{t-1}^{(k)} \gets \boldsymbol{\hat{z}}_{t-1}^{(k-1)} + \zeta \nabla_{\text{meas}}^{(k)}$
    \EndFor
    \State $\boldsymbol{z}_{t-1} \gets \boldsymbol{\hat{z}}_{t-1}^{(K)}$
\EndFor

\State $\hat{\boldsymbol{x}} \gets \mathcal{D}(\boldsymbol{z}_0)$ \Comment{Decode final estimate}
\State \Return $\hat{\boldsymbol{x}}$
\end{algorithmic}
\end{algorithm}

We leverage the diffusion model as a prior to enhance radar point clouds by combining it with radar measurements, as shown in Figure \ref{fig2_method_framework}.(c) and (d). Algorithm \ref{alg1} depicts the process, and Appendix \ref{app_a} depicts the proof of our method. The algorithm begins by initializing a variable $\boldsymbol{z}_T$ with random noise from a Gaussian distribution $\mathcal{N}(\boldsymbol{0}, \boldsymbol{I})$, setting the starting point for the diffusion process. This is typical in diffusion models, which start from pure noise and gradually denoise to generate data. The main loop then iterates from $t=T$ to $t=1$, representing the reverse diffusion process where the state is refined at each step. Within this loop, there are two key updates:

a) \textbf{Diffusion Model Update}: The algorithm calculates $\mu_\delta$, the predicted mean, using a neural network $S_\delta(\boldsymbol{z}_t, t)$. It then updates $\boldsymbol{\hat{z}}_{t-1}$ using the formula $\boldsymbol{z}_t + \lambda_{\text{diff}} \boldsymbol{\Sigma}_\delta(\boldsymbol{z}_t, t)^{-1} (\boldsymbol{\mu}_\delta - \boldsymbol{z}_t)$, where $\lambda_{\text{diff}}$ is the diffusion coefficient and $\boldsymbol{\Sigma}_\delta$ is the covariance matrix. This step moves the state towards the predicted mean, scaled by the inverse covariance, aligning with the diffusion model's goal of denoising.

b) \textbf{Measurement Model Update}: After the diffusion update, there’s an inner loop for $K$ iterations to incorporate measurement data. It starts with $\boldsymbol{\hat{z}}_{t-1}^{(0)}$ and uses Tweedie's formula to compute $\bar{\boldsymbol{z}}_0^{(k)} \gets \frac{1}{\sqrt{\alpha_t}}(\boldsymbol{\hat{z}}_{t-1}^{(k-1)}-\sqrt{1-\alpha_t}S_\delta(\boldsymbol{\hat{z}}_{t-1}^{(k-1)},t))$, which estimates the initial state from the current diffusion state. This is then decoded to $\boldsymbol{x}_0^{(k)}$ using a decoder $D$, mapping the latent space to the data space. The measurement error is computed as $\nabla_{\text{meas}}^{(k)} \gets \nabla_{\boldsymbol{z}} \| \gamma\boldsymbol{Y}-{A}(\boldsymbol{x}_0^{(k)}) \|_2^2$, where the  $\gamma$ is a measurement scale, $\boldsymbol{Y}$ is the measurement, and ${A(\cdot)}$ is the measurement system forward function. The state $\boldsymbol{\hat{z}}_{t-1}^{(K)}$ is updated using this gradient scaled by $\zeta$, the measurement step size, ensuring the estimate aligns with the data. The probabilistic transition difference between our method and vanilla diffusion model is analyzed in Appendix \ref{app_b}.

\section{Experiments and Results}\label{result}

\subsection{Dataset and Evaluation Metrics}

\begin{wraptable}{r}{0.65\textwidth}  
\renewcommand{\arraystretch}{1.25}
\caption{\small \textbf{Parameters of model in different datasets}}
\label{table2:parameters_details}
\scalebox{0.6}{
\begin{tabular}{ccccc}
\hline
                                 &                                                    &                          & RADIal    & K-Radar         \\ \hline
\multirow{8}{*}{Training}  & \multirow{5}{*}{\makecell{First Stage \\ (VQ-VAE)}}               & Original LiDAR Size          & $512\times768\times3$ & -               \\
                                 &                                                    & Encoder/Decoder Layers   & 5         & -               \\
                                 &                                                    & Latent Embedded Dimension & 4         & -               \\
                                 &                                                    & Latent Embedded Channels  & 2048      & -               \\
                                 &                                                    & Training Batch Size      & 2         & -               \\ \cline{2-5} 
                                 & \multirow{3}{*}{\makecell{Second Stage \\ (Latent Diffusion)}}   & Latent LiDAR Size          & $32\times48\times4$   & -               \\
                                 &                                                    & Encoder/Decoder Layers   & 3         & -               \\
                                 &                                                    & Training Batch Size      & 16        & -               \\ \hline
\multirow{5}{*}{Inference} &                                                    & Input Radar Size         & $512\times768\times3$  & $512\times768\times3$       \\
                                 &                                                    & DDIM Steps               & 1000      & 500             \\
                                 &                                                    & Measurement Scale ($\zeta$)       & 1.0       & 1.0             \\
                                 &                                                    & Measurement Step Size ($\gamma$)   & 0.001     & 0.0005          \\
                                 &                                                    & Measurement Step Number ($K$) & 20        & 5               \\ \hline
\end{tabular}
}
\end{wraptable}

\textbf{Dataset}:
To evaluate the effectiveness of our proposed method, two autonomous driving sensing datasets, RADIal dataset \citep{rebut2022raw} and K-Radar dataset \citep{paek2022kradar}, are selected. The RADIal dataset is a collection of 2 hours of raw data from synchronized automotive-grade sensors (camera, laser, High Definition radar) in various environments (city-street, highway, countryside road). It comes with GPS and the vehicle’s CAN traces.
From the dataset, City (30.2\%), countryside (50.0\%) and highway (18.2\%) scenarios are included.
The K-Radar dataset is a large-scale 3D perception benchmark for autonomous driving, featuring 35,000 frames of 4D Radar tensor (4DRT) data with power measurements across Doppler, range, azimuth, and elevation dimensions. In our experiments, the K-Radar dataset is only used for cross dataset validation.



\textbf{Evaluation Metrics}:
Based on previous research \citep{zhang2024towards, wu2024diffradar, luan2024diffusion}, Chamfer Distance (CD) is primarily used in our experiments to evaluate the mutual minimum distance between generated 3D points and ground truth 3D points. This metric assesses the similarity of two point sets, with smaller CD values indicating higher similarity.
In addition to Chamfer Distance, we also consider other metrics such as Unidirectional Chamfer Distance (UCD), Modified Hausdorff Distance (MHD), Unidirectional Modified Hausdorff Distance (UMHD), and Fréchet Inception Distance (FID).

\textbf{Implementation Details}:
Our method trains a diffusion model exclusively on LiDAR data only (e.g., LiDAR point cloud in RADIal dataset), employs the trained model to reconstruct LiDAR style output as prior information during inference procedure, to generate the posterior high resolution radar point cloud by applying the prior on top of likelihood generated by radar heatmap input (e.g., radar heatmap in RADIal, K-Radar or any other dataset). Once trained, the diffusion model which generates the prior can be applied with any other radar data input for inference using our proposed approach.

a) \textbf{Pre-processing}:  
RADIal dataset LiDAR points are first projected into a polar coordinate BEV image with a resolution of \( 512 \times 768 \), covering an angular range of \([-90^\circ, 90^\circ]\) (512 divisions) and a radial range of \([0, 103]\) m (768 divisions). The created LiDAR BEV image is a binary mask, with occupied grids set to 1 and empty grids to 0. For radar data, the RADIal dataset API decodes the range-azimuth heatmap from the raw signal matrix into a \( 512 \times 768 \) image, with values normalized to \([0, 1]\) via linear scaling. Similarly, the K-Radar dataset's range-azimuth signals are used, with a Field-of-View from \(-53^\circ\) to \( 53^\circ\) in azimuth and a detection range of \([0, 118]\) m. The K-Radar data is interpolated from its original \( 256 \times 128 \) size to match the \( 512 \times 768 \) resolution of the RADIal dataset. No additional training is required for K-Radar dataset.

b) \textbf{Model Training}:  
The training process begins with a diffusion model trained exclusively on LiDAR data of RADIal dataset, serving as the prior. Two stages are included. 
The first stage is training a VQ-VAE on an Nvidia L40 GPU with 48 GB of CUDA memory. The training parameters are detailed in Table \ref{table2:parameters_details}. The LiDAR BEV images are duplicated into 3 channels to give full play to the feature capture capability of VAE. 
Once the VQ-VAE converges after 62 epochs, the second stage—training the latent diffusion model—commences. This model reaches convergence after 130 epochs.  

c) \textbf{Model Inference}:  
For RADIal dataset, radar data is used as input measurement $\boldsymbol{Y}$, refining the $32 \times 48 \times 4$ latent variable $\boldsymbol{z}$ together with diffusion prior gradient. The final output is obtained after 1000 iterative steps and decoding $\boldsymbol{z}$. Table \ref{table2:parameters_details} specifies the parameter values for the RADIal dataset: $\zeta = 1.0$, $\gamma = 0.001$, and $K = 20$.  
For the K-Radar dataset, $\zeta = 1.0$, $\gamma = 0.0005$, and $K = 5$. The previous diffusion prior is used repetitively. A deep manifold searching is needed to acquire the best parameters, which is analyzed in Appendix \ref{app_d}.

d) \textbf{Post-processing}: Once the output has converged, an average is computed across the three channels, which suppresses $512\times768\times3$ to $512\times768$. A threshold of 0.01 is then applied to identify valid BEV grid points for generating the final radar points. After extracting the grid, the output is converted from polar to Cartesian coordinates.

\subsection{Results}

\subsubsection{Performance Evaluation of RADIal Dataset}

Statistical performance analysis on the RADIal dataset can demonstrate the effectiveness of the proposed method. 
Table \ref{table_1_enhancement_quantity_analys} presents the performance metrics of point cloud enhancement results for different methods.
Five training methods, including supervised and unsupervised, are selected to compare the performance with our proposed method. Unsupervised methods include CFAR, $L_1$ Reg and $L_2$ Reg. CFAR is a traditional peak extraction method. The other two are methods that originally solved radar angle estimation as an inverse problem.
RadarHD and Diffradar are two typical supervised approaches that have recently been proposed. The first is a discriminative model using UNet \citep{UNet}, while the second is a generative model which is based on conditional-DDPM \citep{ho2020denoising}. Notably, there are many works in each category, but only RadarHD is open source. We reproduced Diffradar's data.

\vspace{-5pt}
\renewcommand\arraystretch{1.0}
\begin{table}[h]
\caption{\textbf{Statistical analysis of point enhancement on RADIal datasets.} Our method outperforms other unsupervised approaches and achieves performance comparable to supervised methods.}
\begin{center}
\scalebox{0.75}{

\label{table_1_enhancement_quantity_analys}
\begin{tabular}{ccccccc}
\hline
Methods   & Supervised Training& $\text{FID}_{BEV}$ $\downarrow$ & CD $\downarrow$  & UCD $\downarrow$  & MHD $\downarrow$    & UMHD $\downarrow$  \\ \hline
CFAR     & \ding{55}  & 242.17   & 6.38 & 3.11 & 2.06 & 1.94 \\ 
$L_1$ Reg \citep{shkvarko2016solving}  & \ding{55}   & 576.09   & 10.24 & 6.55 & 5.34 & 5.28 \\ 
$L_2$ Reg \citep{shkvarko2016solving} & \ding{55}  & 553.60   & 10.40 & 6.41 & 5.50 & 5.34 \\ \hline
RadarHD \citep{prabhakara2023high}   & \checkmark  & \greenuline{174.90}   & 6.50 & \textbf{1.87} & 3.05 & 1.47\\ 
Diffradar \citep{wu2024diffradar} & \checkmark  & \textbf{138.07}   & \textbf{5.64} & \greenuline{2.25} & \textbf{1.60} & \textbf{0.97} \\ \hline
Ours   & \ding{55} & 272.67   & \greenuline{6.31} & 3.70 & \greenuline{1.61} & \greenuline{1.23} \\ 
\hline
\end{tabular}
}
\end{center}
\end{table}

\begin{figure}[h]
    \centering
    \includegraphics[width=1\linewidth]{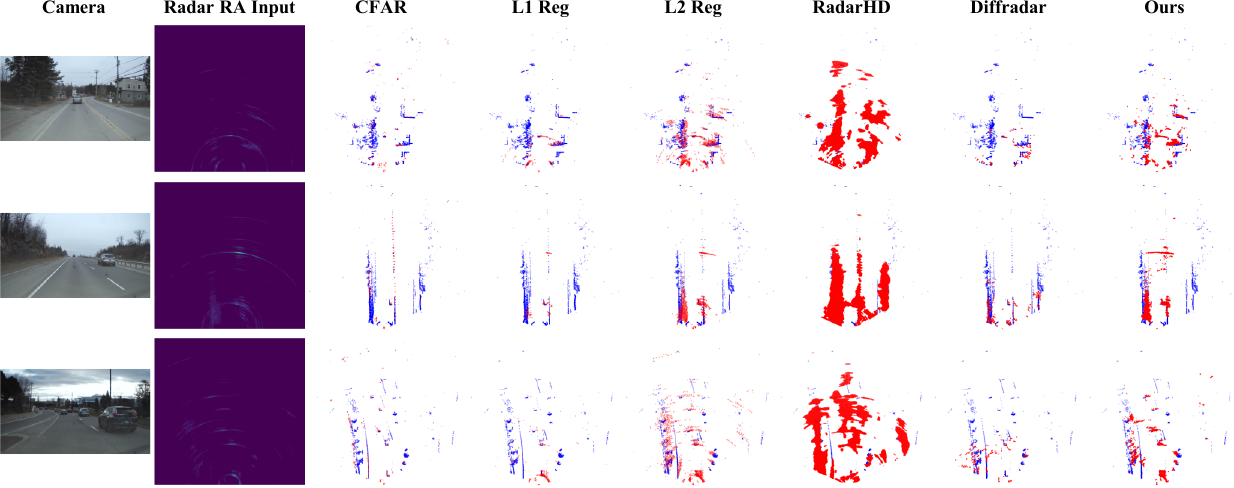}
    \caption{\textbf{Qualitative comparison of the RADIal dataset across different methods}. Different randomly selected frames from three scenarios are displayed in rows, while point enhancement results from different methods are shown in columns. In each sub-figure, \textcolor{blue}{blue} points represent the LiDAR ground truth, while \textcolor{red}{red} points indicate the enhanced point cloud outputs. More results please refer to Figure \ref{fig:appendix_vis_radial} in Appendix \ref{app_g}. 
     \vspace{-15pt}} 
    \label{fig:final_results_all}
\end{figure}

%
It can be observed that our method achieves a comparable level of performance to the supervised learning method, RadarHD.
Compared to Diffradar, the point clouds generated by our method exhibit good similarity and density.
Figure \ref{fig:final_results_all} shows the results of various radar super-resolution point cloud generation schemes. 
Compared to the sparse results of CFAR, other schemes provide denser point clouds. However, the $L_1$/$L_2$ regularization schemes introduce a significant number of noise points. 
For the RadarHD, the main issue is the complete loss of generated details.
For Diffradar, while the model indeed has high quality point cloud generation capabilities, it suffers from deficiencies in density. 
In contrast, our method generates point clouds with higher density than the CFAR method, lower noise levels than the $L_1$/$L_2$ regularization, more detailed point clouds than RadarHD, and higher fidelity than Diffradar.
In addition, the inference variance and acceleration methods are explored in Appendix \ref{app_e} and \ref{app_f}.

\subsubsection{Cross-Dataset Validation on K-Radar Dataset}

\begin{figure}[h]
    \centering
    \includegraphics[width=1.0\linewidth]{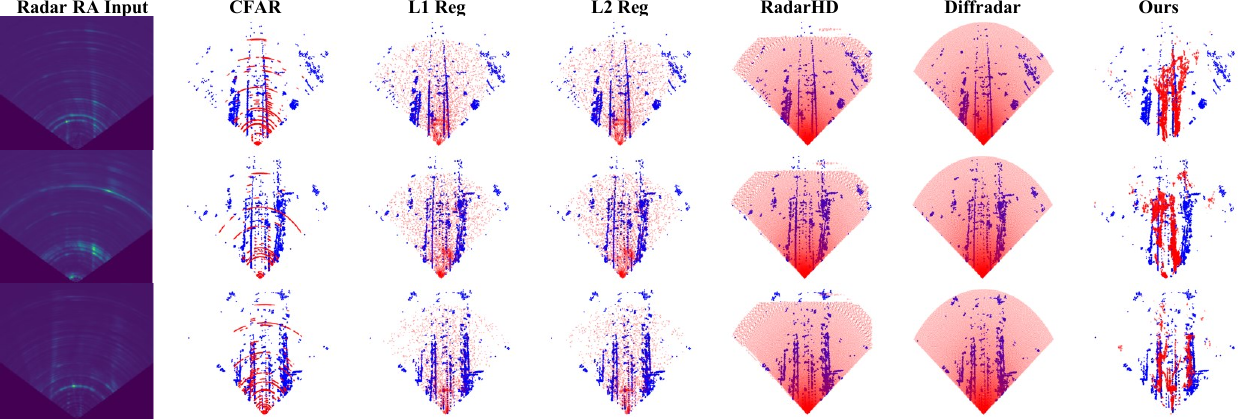}
    \caption{\textbf{Cross-dataset enhancement on K-Radar}. Rows show radar inputs; columns show outputs from different methods. \textcolor{blue}{Blue} and \textcolor{red}{red} points denote LiDAR (GT) and generated points respectively. Other methods struggle with noise and domain shifts, while our approach yields cleaner, robust results without requiring paired training data. More results please refer to Figure \ref{fig:appendix_vis_kradar} in Appendix \ref{app_g}.
    \vspace{-15pt}}
    \label{fig9:kradar_recon_results}
\end{figure}

To illustrate the generalization ability of our method, we directly use radar data from K-Radar dataset as input for each method; no more LiDAR data from K-Radar dataset is included. The inference parameters are listed in Table \ref{table2:parameters_details}. The statistical and visualization results are depicted in Table \ref{table3: kradar_performance} and Figure \ref{fig9:kradar_recon_results} respectively. 

\vspace{-11pt}
\begin{table}[h]
\captionsetup{skip=10pt} 
\caption{\textbf{Cross dataset validation on K-Radar dataset}. Our method demonstrates strong generalizability compared to other approaches.}
\label{table3: kradar_performance}
\centering
\scalebox{0.75}{
\begin{tabular}{ccccccc}
\hline
Method    & Supervised Training & $\text{FID}_{BEV}$  $\downarrow$           & CD  $\downarrow$           & UCD  $\downarrow$         & MHD  $\downarrow$          & UMHD  $\downarrow$          \\ \hline
CFAR      &\ding{55} & \greenuline{244.55}          & \greenuline{26.09}          & \greenuline{13.27}         & \greenuline{22.82}          & \textbf{135.53} \\
$L_1$ Reg \citep{shkvarko2016solving} &\ding{55}    & 263.17          & 49.84          & 26.99         & 46.99          & 386.02          \\
$L_2$ Reg \citep{shkvarko2016solving} &\ding{55}  & 258.69          & 33.34          & 26.94         & 42.46          & 386.60          \\ \hline
RadarHD \citep{prabhakara2023high} &\checkmark & -          & -          & -         & -          & -          \\
Diffradar  \citep{wu2024diffradar} &\checkmark & -          & -          & -         & -          & -          \\ \hline
Ours     &\ding{55} & \textbf{230.67} & \textbf{23.42} & \textbf{7.38} & \textbf{16.94} & \greenuline{361.91}          \\ \hline
\end{tabular}
}
\end{table}

It is evident that supervised methods like RadarHD and DiffRadar produce overly dense outputs that saturate the detection space, highlighting their limitations in handling cross-dataset generalization. Similarly, $L_1$/$L_2$ regression-based methods yield noisy results, while traditional CFAR approaches produce sparse outputs with side-lobe noise. In contrast, our method more accurately recovers the basic structure of the environment, despite missing some regions detected by LiDAR. For instance, as shown in the first row of the figure, points along the left roadside edge and at farther distances are absent due to weak signals in the radar input and occlusion caused by the radar's low mounting position on the K-Radar vehicle. Additionally, recovery is constrained by our use of only 16-layer LiDAR as a prior, while K-Radar employs a higher-resolution 64-layer LiDAR, limiting the achievable output fidelity.

\begin{table}[h]
\caption{\textbf{Cross scenario validation on RADIal dataset using CD metrics}. The diffusion prior is trained under LiDAR data from Countryside and All of RADIal dataset. Then radar data from different scenarios (City, Countryside and Highway) is used as our model's input.\vspace{10pt}}
\label{table4:cross_scene_performance}
\centering
\scalebox{0.75}{
\begin{tabular}{ccccc}
\hline
\multirow{2}{*}{Prior} & \multicolumn{4}{c}{Test Scenario (CD)}           \\ \cline{2-5} 
                       & City (30.2\%) & Countryside (50.0\%) & Highway (18.8\%) & All \\ \hline
Countryside            & 7.46    & \textbf{5.85}           & 7.48       & 6.59   \\
All                    & \textbf{6.77}    & 6.26           & \textbf{6.01}       & \textbf{6.31}   \\ \hline
\end{tabular}}
\vspace{-10pt}
\end{table}

\subsubsection{Cross Scenario Validation on RADIal Dataset}
\label{cross_scen_valid}

\begin{wrapfigure}{r}{0.55\textwidth}
    \centering
    \vspace{-10pt} 
    \includegraphics[width=0.5\textwidth]{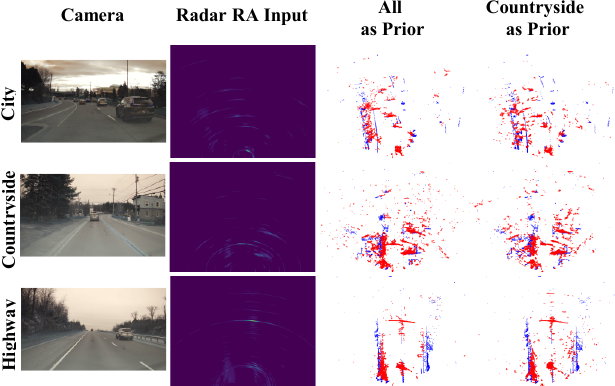}
    \caption{\textbf{Visualization results of cross-scenario validation on RADIal dataset}. Three rows demonstrate three scenarios provided by RADIal dataset. In each scenario, two frames are selected. Results of all as prior and countryside as prior are provided. More results please refer to Figure \ref{fig:appendix_vis_cross_scenario} in Appendix \ref{app_g}. \vspace{-10pt}}
    \label{fig10:cross_scenario}
\end{wrapfigure}

The generalization capability of our method is validated through cross-scenario testing on the RADIal dataset, which includes three scenarios: City (30.2\%), Countryside (50\%), and Highway (18.8\%). Table \ref{table4:cross_scene_performance} presents the cross-scenario performance using CD metrics. Training the diffusion prior with all RADIal LiDAR data yields better overall performance across scenarios (CD: 6.31). However, when training solely on Countryside LiDAR data, the model achieves superior performance in the Countryside scenario (CD: 5.85) but underperforms in City and Highway scenarios, leading to a higher overall CD (6.59). The reason is that the diffusion prior is specifically tailored to the Countryside LiDAR data distribution, capturing its unique spatial patterns more effectively. Despite this, the model’s generalization ensures effective point cloud enhancement even without prior data from City and Highway scenarios.

Figure \ref{fig10:cross_scenario} provides a visual comparison of output results across various data subsets. When relying solely on countryside LiDAR data as a prior, the enhanced radar point clouds show increased noise in City and Highway settings, but superior performance in the Countryside itself. Among the different scenarios, the highway environment delivers the strongest performance owing to its simpler layout, whereas the city scenario yields somewhat less precise results due to its intricate urban landscape and varied object distributions, which significantly affect radar input signals.

\section{Conclusion and Future Work}\label{conclusion}

This paper proposes a radar point cloud enhancement algorithm using a diffusion model as a prior.
Bayesian inference for the radar enhancement is formulated, and the samples are drawn from the posterior distribution using the diffusion model as a prior and the radar imaging equation as a constraint.
Through parameter analysis and comparative experiments, the effectiveness and high performance of our method are demonstrated. 
However, challenges such as the high cost of inference time and the alignment of multi-modal features persist, and will be examined and addressed in future work.

{
\small



\bibliographystyle{unsrt}
\bibliography{references}

}

\newpage

\appendix


\section{Proof of Angle Estimation Using a Diffusion Model Prior} \label{app_a}
To solve the problem depicted
in Eq. (\ref{bayesian_theorem}) with a diffusion model, the posterior distribution is expressed as:
\begin{equation}
    p(\boldsymbol{x}|\boldsymbol{Y}) \propto p(\boldsymbol{Y}|\boldsymbol{x}) p_\delta(\boldsymbol{x}).
\end{equation}
To reduce GPU memory usage, a latent diffusion model is applied, reformulating the problem in the latent space:
\begin{equation}
    p(\boldsymbol{z}|\boldsymbol{Y}) \propto p(\boldsymbol{Y}|\mathcal{D}(\boldsymbol{z})) p_\delta(\boldsymbol{z}),
\end{equation}
where \( \boldsymbol{z} \) is the latent representation of \( \boldsymbol{x} \), and \( \mathcal{D}(\cdot) \) is the decoder. During each iteration, the update follows:
\begin{equation}
\label{eq22}
    p(\boldsymbol{z}_{t-1}|\boldsymbol{z}_t, \boldsymbol{Y}) \propto p(\boldsymbol{Y}|\mathcal{D}(\hat{\boldsymbol{z}}_0)) p_\delta(\boldsymbol{z}_{t-1}|\boldsymbol{z}_t),
\end{equation}
where \( \hat{\boldsymbol{z}}_0 \) is an estimate of \( \boldsymbol{z}_0 \), since \( \boldsymbol{x}_0 = \mathcal{D}(\boldsymbol{z}_0) \), but \( \boldsymbol{x}_t \neq \mathcal{D}(\boldsymbol{z}_t) \). Thus, the measurement model gradient must be computed using \( \hat{\boldsymbol{z}}_0 \).

To maximize Eq. (\ref{eq22}), we apply Maximum A Posteriori (MAP) estimation. The logarithmic posterior is:
\begin{equation}
\label{MAP_eq}
    \log p(\boldsymbol{z}_{t-1}|\boldsymbol{z}_t, \boldsymbol{Y}) \propto \log p(\boldsymbol{Y}|\mathcal{D}(\hat{\boldsymbol{z}}_0)) + \log p_\delta(\boldsymbol{z}_{t-1}|\boldsymbol{z}_t).
\end{equation}
Differentiating and setting to zero gives:
\begin{equation}
\label{MLE_solve_joint_prob}
    \nabla_{\boldsymbol{z}_{t-1}} \log p(\boldsymbol{z}_{t-1}|\boldsymbol{z}_t, \boldsymbol{Y}) = 0,
\end{equation}
yielding:
\begin{equation}
\label{measurement_gradient}
    \nabla_{\boldsymbol{z}_{t-1}} \log p(\boldsymbol{Y}|\mathcal{D}(\hat{\boldsymbol{z}}_0)) + \nabla_{\boldsymbol{z}_{t-1}} \log p_\delta(\boldsymbol{z}_{t-1}|\boldsymbol{z}_t) = 0.
\end{equation}

The diffusion model gradient is:
\begin{align}
\label{diffusion_grad}
    \nabla_{\boldsymbol{z}_{t-1}} \log p_\delta(\boldsymbol{z}_{t-1}|\boldsymbol{z}_t) &= \nabla_{\boldsymbol{z}_{t-1}} \left( -\frac{1}{2} (\boldsymbol{z}_{t-1} - \boldsymbol{\mu}_\delta(\boldsymbol{z}_t, t))^\top \boldsymbol{\Sigma}_\delta(\boldsymbol{z}_t, t)^{-1} (\boldsymbol{z}_{t-1} - \boldsymbol{\mu}_\delta(\boldsymbol{z}_t, t)) \right) \notag \\
    &= -\boldsymbol{\Sigma}_\delta(\boldsymbol{z}_t, t)^{-1} (\boldsymbol{z}_{t-1} - \boldsymbol{\mu}_\delta(\boldsymbol{z}_t, t)),
\end{align}
where \( \boldsymbol{\mu}_\delta(\boldsymbol{z}_t, t) \) is predicted by the neural network $\epsilon_\delta (\cdot)$. The measurement model gradient is:
\begin{equation}
\label{gradient_of_meas}
    \nabla_{\boldsymbol{z}_{t-1}} \log p(\boldsymbol{Y}|\mathcal{D}(\hat{\boldsymbol{z}}_0)) = \nabla_{\boldsymbol{z}_{t-1}} \| \boldsymbol{Y} - {A}(\mathcal{D}(\hat{\boldsymbol{z}}_0)) \|_2^2.
\end{equation}
Since only \( \boldsymbol{z}_t \) is available at step \( t \), Tweedie’s formula \citep{song2023solving} estimates \( \hat{\boldsymbol{z}}_0 \):
\begin{equation}
    \hat{\boldsymbol{z}}_0 = \frac{1}{\sqrt{\alpha_t}} \left( \boldsymbol{z}_t - \sqrt{1 - \alpha_t} S_\delta(\boldsymbol{z}_t, t) \right).
\end{equation}

The diffusion prior gradient ensures the output aligns with the LiDAR data domain, while the measurement gradient regulates fidelity to the radar measurement \( \boldsymbol{Y} \). These gradients are iteratively applied to an initial noise input, with step and weight parameters optimized for performance, as detailed in Algorithm \ref{alg1}.
%

\section{Likelihood Function of Parallel Radar Imaging} \label{app_c}
The measurement forward function of Eq. \ref{radar_measurement_function} can be described in details as:
\small
\begin{align}
    \boldsymbol{Y} 
    &= \left[\boldsymbol{s}_1, \boldsymbol{s}_2, \cdots, \boldsymbol{s}_M\right]  + \boldsymbol{H} = 
    \begin{bmatrix}
        0 & 0 & \cdots & 0 \\
        \frac{d\cos(\theta_1)}{c} & \frac{d\cos(\theta_2)}{c} & \cdots & \frac{d\cos(\theta_M)}{c} \\
        2\frac{d\cos(\theta_1)}{c} & 2\frac{d\cos(\theta_2)}{c} & \cdots & 2\frac{d\cos(\theta_M)}{c} \\
        \vdots & \vdots & \ddots & \vdots \\
        \left(N-1\right)\frac{d\cos(\theta_1)}{c} & \left(N-1\right)\frac{d\cos(\theta_2)}{c} & \cdots & \left(N-1\right)\frac{d\cos(\theta_M)}{c}
    \end{bmatrix}
     + \boldsymbol{H}
    .
\end{align}

\begin{figure}[h]
    \centering
    \includegraphics[width=1.0\linewidth]{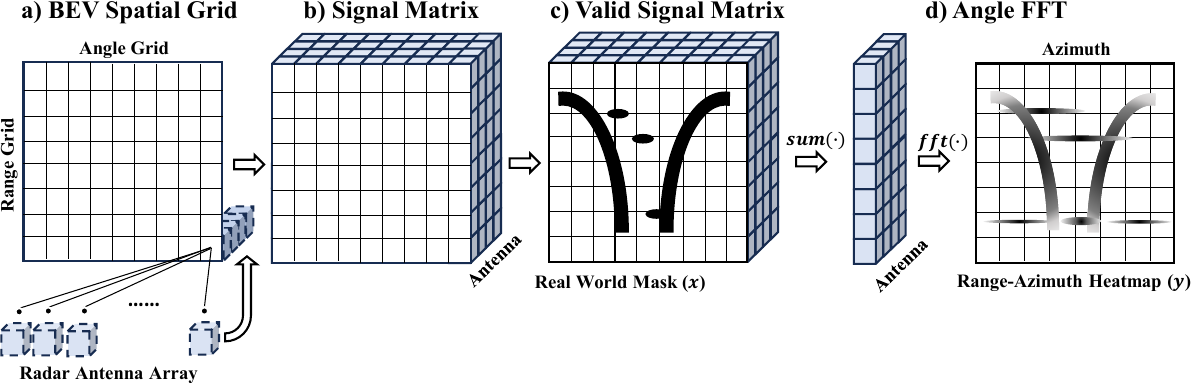}
    \caption{Parallel radar imaging likelihood function data process.}
    \label{fig7:parallel_radar_imaging}
\end{figure}

To compute the gradient from measurement model $A(\cdot)$ as Eq. (\ref{gradient_of_meas}) depicts, this measurement function must be differentiable. However, in real implementation, this step is usually discrete, which will break PyTorch computation graph. Therefore, a trick is applied here to ensure the computable of our method. The key idea is assuming real world $\boldsymbol{x}$ under bird eye's view (BEV) as a mask. Then use this mask to filter the spatial signal on radar antenna array. This process can be described as Figure \ref{fig7:parallel_radar_imaging}.

At the step a), a BEV spatial grid is initialized as the target sensing space for radar sensor. The space is under polar coordinate. Suppose there is an object in each spatial grid that can reflect microwave electromagnetic wave to the radar sensor, the signal on each radar antenna is computable theoretically. Then in b), for each spatial grid, a signal matrix that consisted of range, angle and antenna is formed. However, the signal matrix represents full space signal, which is impossible because the target space could not be full filled with objects. Therefore, the real world mask $x$ is used to filter out valid signal grid in c). Then in step d), after summation of the complex signal from different valid grids on the same antenna, the range-antenna signal matrix is constructed. Finally, after angle FFT, the range-azimuth heatmap $\boldsymbol{Y}$ result of radar imaging process is given.

\begin{figure}[h]
    \centering
    \includegraphics[width=1.0\linewidth]{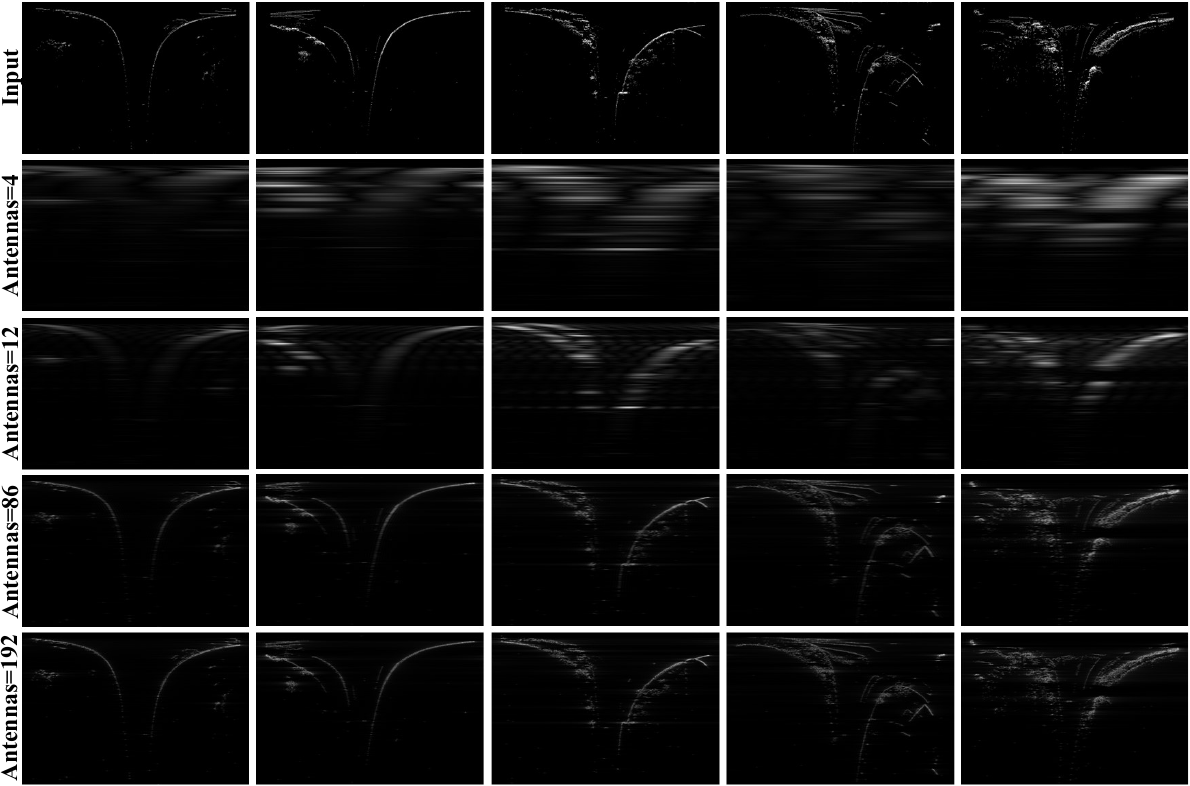}
    \caption{Experiment results of parallel radar imaging. Results of antennas number of 4, 12, 86 and 192 are showed in rows with different inputs in columns.}
    \label{fig8:parallel_imging_exp}
\end{figure}

To illustrate the effectiveness of the parallel radar imaging process, the LiDAR data is used as input $x$. The results are showed in Figure \ref{fig8:parallel_imging_exp}. Different antennas number setting corresponding to different radar hardware setting. Antennas=4, 12 represent the 1T4R and 3T4R radar respectively. Antennas=86 represents ``TI MMWAVE-2243 CASCADE'' types of radar, while Antennas=192 represents the ideal 12T16R radar. Here ``T'' and ``R'' represent Transmitter and Receiver.

From the visualization results in Figure \ref{fig8:parallel_imging_exp}, the fact is known that the radar imaging function is able to degrade real world $\boldsymbol{x}$. Less number of antennas means more degradation. Therefore, when solving the radar point cloud enhancement inverse problem based on this likelihood function, one possible solution as $\boldsymbol{x}$ that hidden in this function is available.

%
However, the ill-pose issue in the likelihood function limits the capability to totally recover to the $x$ because of too much information loss. Therefore, the diffusion model based regularization method proposed in this paper is needed.


\section{Hyper Parameters Manifold Searching} \label{app_d}

\begin{figure}[ht]
    \centering
    \includegraphics[width=0.9\linewidth]{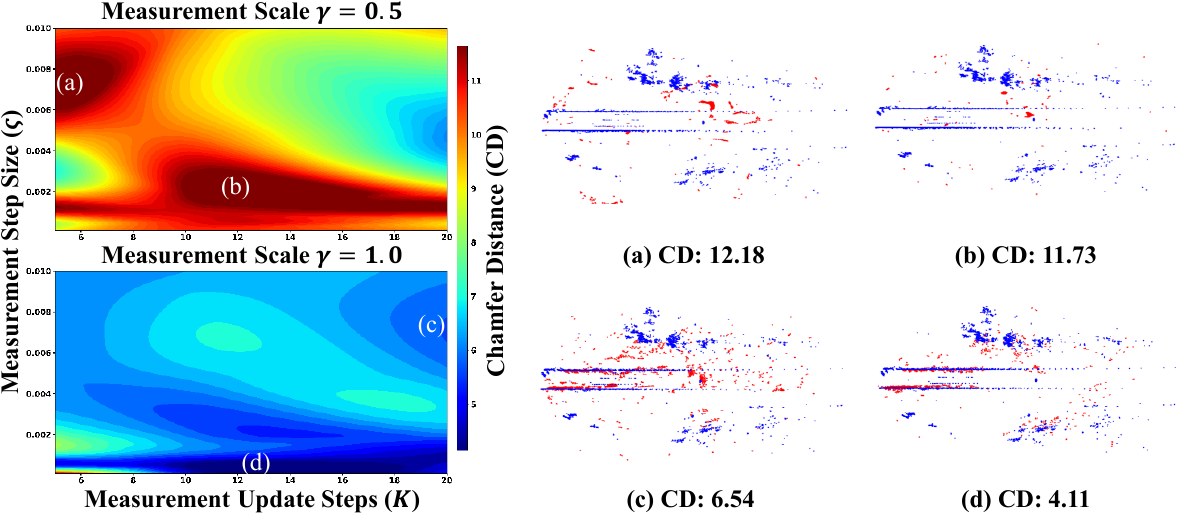}
    \caption{Hyperparameter space manifold analysis of proposed method. The two sub-figures represent model generation results CD metric from measurement scale 0.5 and 1.0. Both of them x-axis represent measurement step number $K$, y-axis represents step size $\zeta$. The output radar points (\textcolor{red}{red}) and ground truth LiDAR points (\textcolor{blue}{blue}) are overlaid from (a) to (d) to show their similarity.}
    \label{fig5: params manifold}
\end{figure}

In our method, there are three hyper-parameters need to be discussed according to the Algorithm \ref{alg1}: Step size ($\zeta$) of measurement model, Step number ($K$) of measurement model, and Measurement scale ($\gamma$).
Empirically, above parameters are set as follows: $\zeta \in [0, 0.01]$, $K \in [5,20]$ and $\gamma \in [0.5,1.0]$.
The hyper parameter space manifold is depicted in Figure \ref{fig5: params manifold}.

From this figure, we can observe multiple local optima in the parameter manifold space. Since a smaller CD value indicates that the generated radar point cloud is closer to the ground truth, the overall best performance is achieved at the parameter combination $(d)$.
For the manifold space slice with a measurement scale ($\gamma$) of 0.5, more measurement steps are needed to compensate for the insufficient regression contribution of each measurement step.
For the manifold space slice with a measurement scale of 1.0, larger values of $K$ and $\zeta$ do not yield better results. Instead, the local optimal parameter combinations are found within $K \in [9,18]$ and $\zeta \in [0,0.001]$.
This is because excessively large or small values of \(K\) or \(\zeta\) will guide the model to output results that are either too close to or too far from the radar input.
If the output is too close to input, the points enhancement effect is poor, while if it is too far from input, the output becomes uncontrolled.
From the visualized parameter combinations (a) to (d) in Figure \ref{fig5: params manifold}, we can see that the outputs of (a) and (b) are overly sparse and unrelated to the radar input.
Starting from (c), the output shape begins to converge, and noise gradually decreases. Finally, it converges to the global optima at (d), where the CD value is reduced to minima $4.11$.

\section{Probabilistic Flow Analysis} \label{app_b}
\begin{figure}[h]
    \centering
    \includegraphics[width=1.0\linewidth]{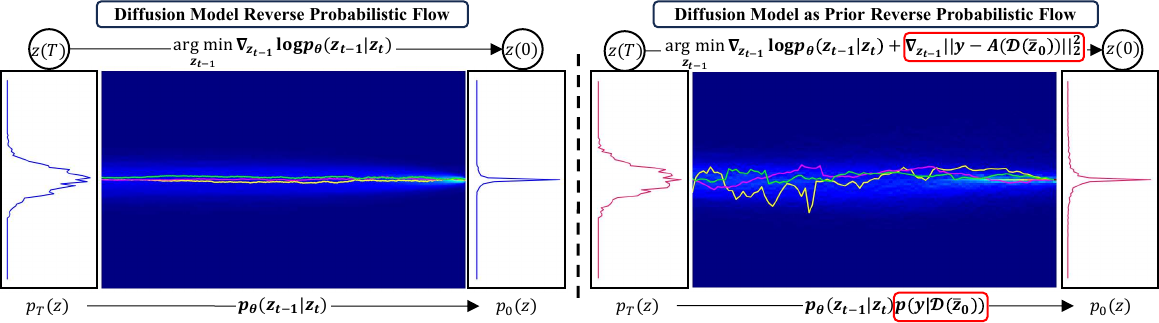}
    \caption{Comparison of reverse probabilistic flow between merely diffusion model (left) and our model using diffusion model as prior (right).}
    \label{fig: probabilistic flow}
\end{figure}

The theoretical foundation of our approach can be understood through the probability flow analysis shown in Figure \ref{fig: probabilistic flow}:

\textbf{Left sub-figure}:
The trained diffusion model demonstrates its capability to transform data from a standard Gaussian distribution $p_T(z)$ back to the target LiDAR distribution $p_0(z)$. This process establishes a learned prior that captures the inherent structure and patterns of LiDAR point clouds.

\textbf{Right sub-figure}:
When incorporating radar measurements as conditions, the 
probability flow follows a modified path while still converging to 
the same target distribution $p_0(z)$. This conditional process ensures: 1) Distribution alignment between enhanced radar points and LiDAR data; 2) Preservation of radar measurement information; and 3) Generalization capability through distribution-level learning.

The similarity in final distributions between these two processes, combined with the measurement-guided optimization, enables our method to generate enhanced point clouds that maintain both LiDAR-like quality and radar measurement fidelity. This distribution-based approach naturally supports generalization across different scenarios, as it learns the underlying data structure rather than direct mappings.

\section{Inference Variation Analysis} \label{app_e}

\begin{figure}[h]
    \centering
    \includegraphics[width=0.6\linewidth]{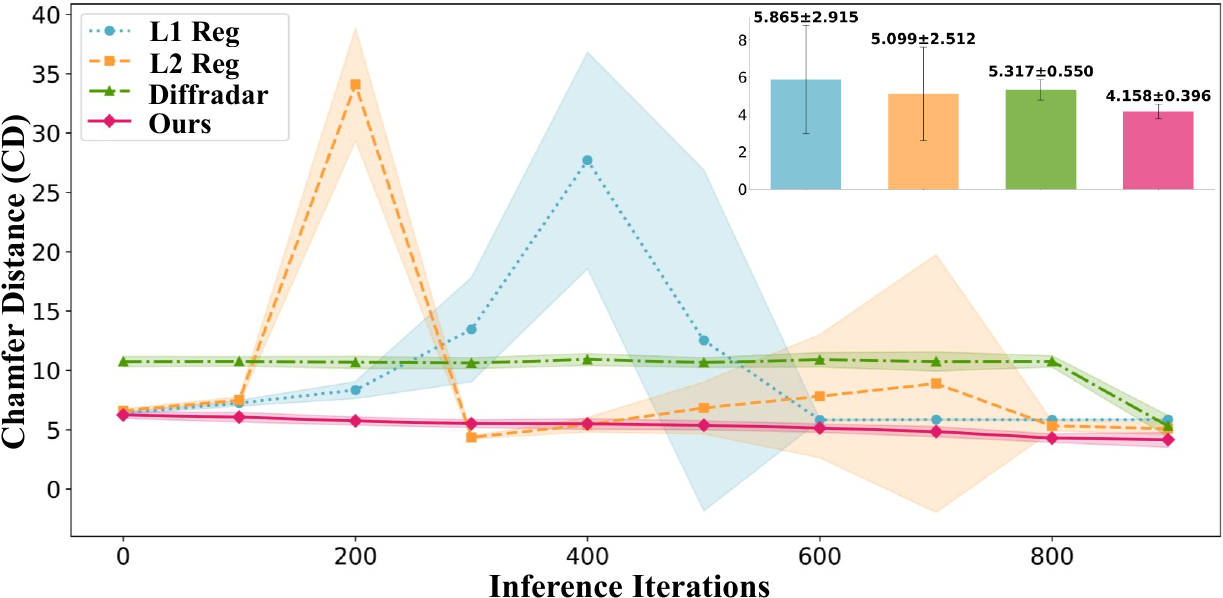}
    \caption{Inference process of radar super-resolution reconstruction using different models. The x-axis shows the inference iterations ranging from 0 to 1000, while the y-axis indicates the Chamfer Distance (CD) between the generated points and the ground truth LiDAR points. The dotted lines represent the mean CD values for each method, and the colored area on each line depict the variation in CD. Up right corner sub-figure depicts the CD mean and variance value at the end iteration.}
    \label{fig3: Inference Variance}
\end{figure}

In the framework of iterative optimization methods for solving inverse problems, the solution space typically contains multiple local minima, and the solutions are significantly influenced by random initial values. Therefore, analyzing the variance of the solutions can effectively reflect the stability of the method.
As illustrated in Figure \ref{fig3: Inference Variance}, four optimization-based methods are discussed: $L_1$/$L_2$ Regularization \citep{shkvarko2016solving}, Diffradar \citep{wu2024diffradar}, and our proposed method.
Five times inference of a same frame radar input are taken to calculate the mean and variance of these methods.
It can be observed that, throughout the inference process, our method reached a lower CD value after the whole inference process.
In addition, our method maintained a low variance during the whole inference process together with lower CD values.

\section{Inference Acceleration}\label{app_f}

\begin{figure}[h]
    \centering
    \includegraphics[width=0.8\linewidth]{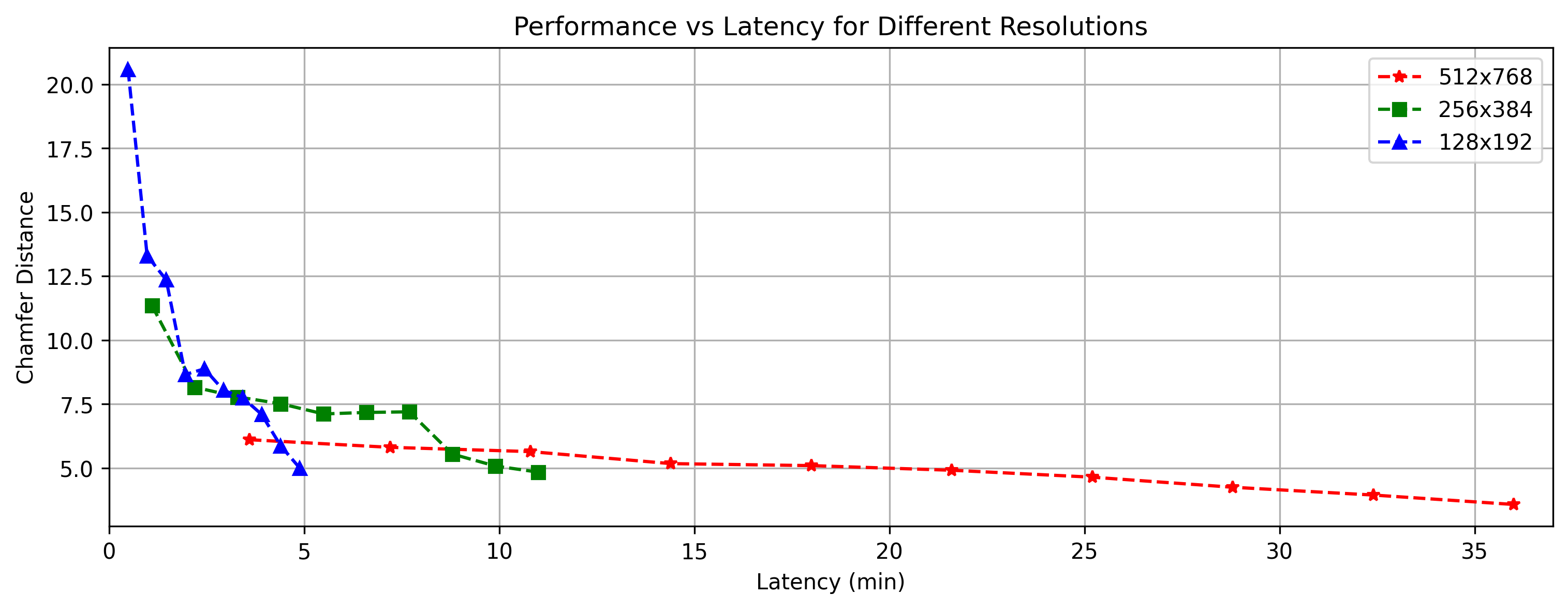}
    \caption{Acceleration of inference process of our proposed method.}
    \label{fig10:accelerate_exp}
\end{figure}

As a computational method, time consumption is a huge burden of current method. There are two ways to accelerate this computation: a). Cut down the input size, b). Early stop. Therefore, a acceleration experiment is implemented and the results are depicted in Figure \ref{fig10:accelerate_exp}.

\begin{figure}[ht]
    \centering
    \includegraphics[width=1.0\linewidth]{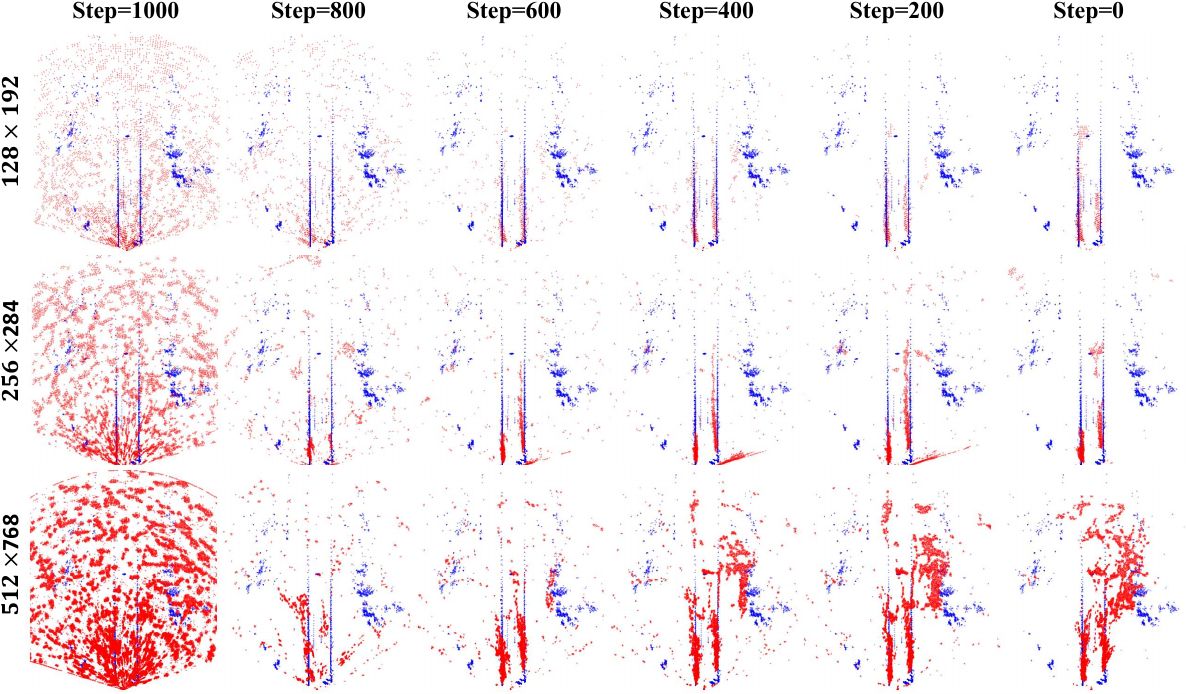}
    \caption{Model performance convergence steps in different input size.}
    \label{fig11:different_scale_input}
\end{figure}

In this experiment, the results of 3 different scales of input data are depicted. In order to maintain the capability of latent features description from VQ-VAE, the feature size of latent feature keeps $32\times48\times4$. From Figure \ref{fig10:accelerate_exp} can we know that, as the input data is down-sampled, inference latency is shorten from 36 minutes to 5 minutes. However, the performance also decreases because of input information loss. 
This decreasing performance visualization can also be observed from Figure \ref{fig11:different_scale_input}. The output points visual density increases as the input data scale increases. In addition, it can be observed that when model early stops at step=600, the visual performance basically approaching ground truth. Therefore, the model's inference latency is 0.4 times of original latency. Combine these two methods, the inference time is shorten to 2 minutes, which is 5.5\% of original time consumption.

Even though with above acceleration approaches, these methods are still not able to meet real-time requirement. However, this section provides two research directions to solve the latency issue, which is one of our works in the future.

\section{More Visualization Results}\label{app_g}
\begin{figure}
    \centering
    \includegraphics[width=1.0\linewidth]{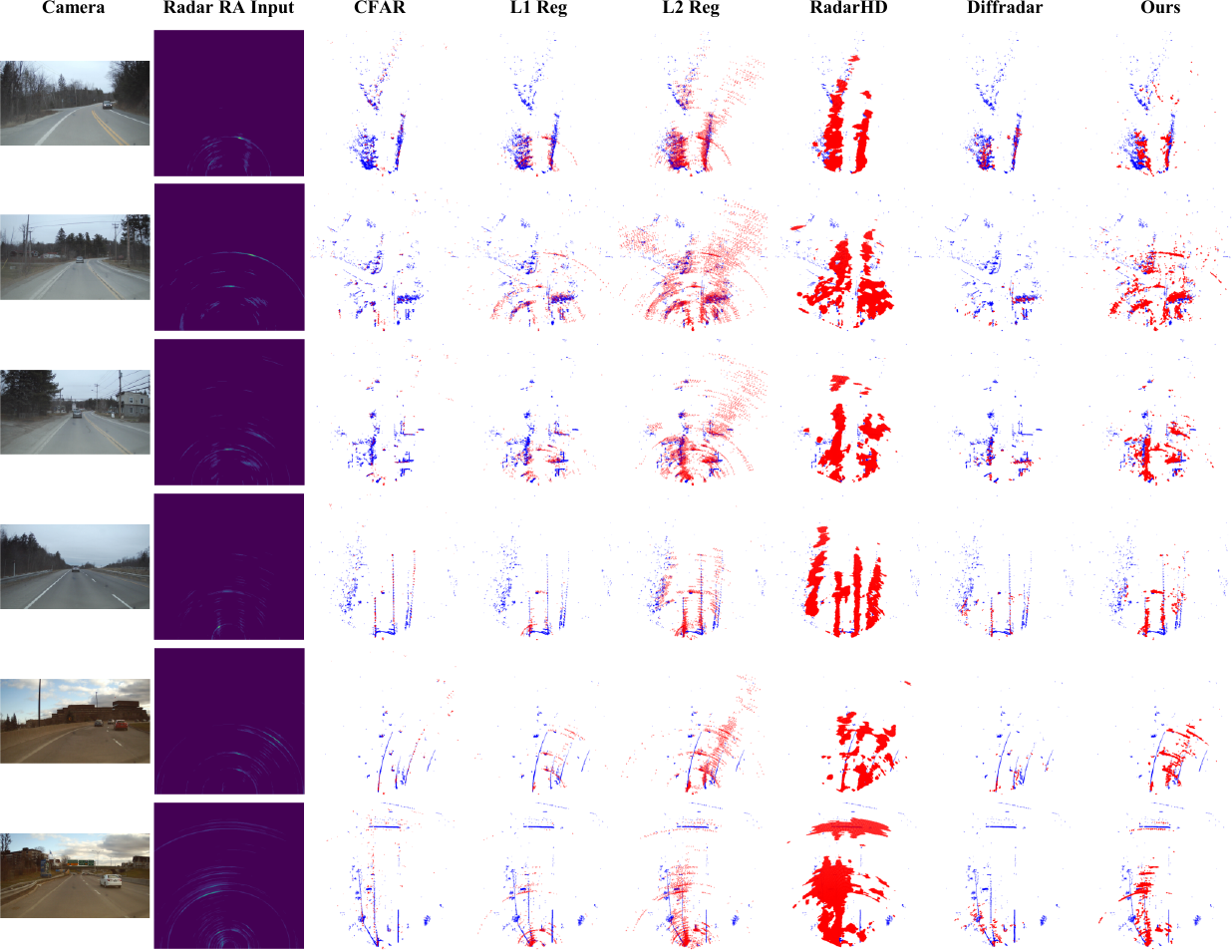}
    \caption{Comparison of radar point cloud enhancement of different methods on RADIal dataset.}
    \label{fig:appendix_vis_radial}
\end{figure}

\begin{figure}
    \centering
    \includegraphics[width=1.0\linewidth]{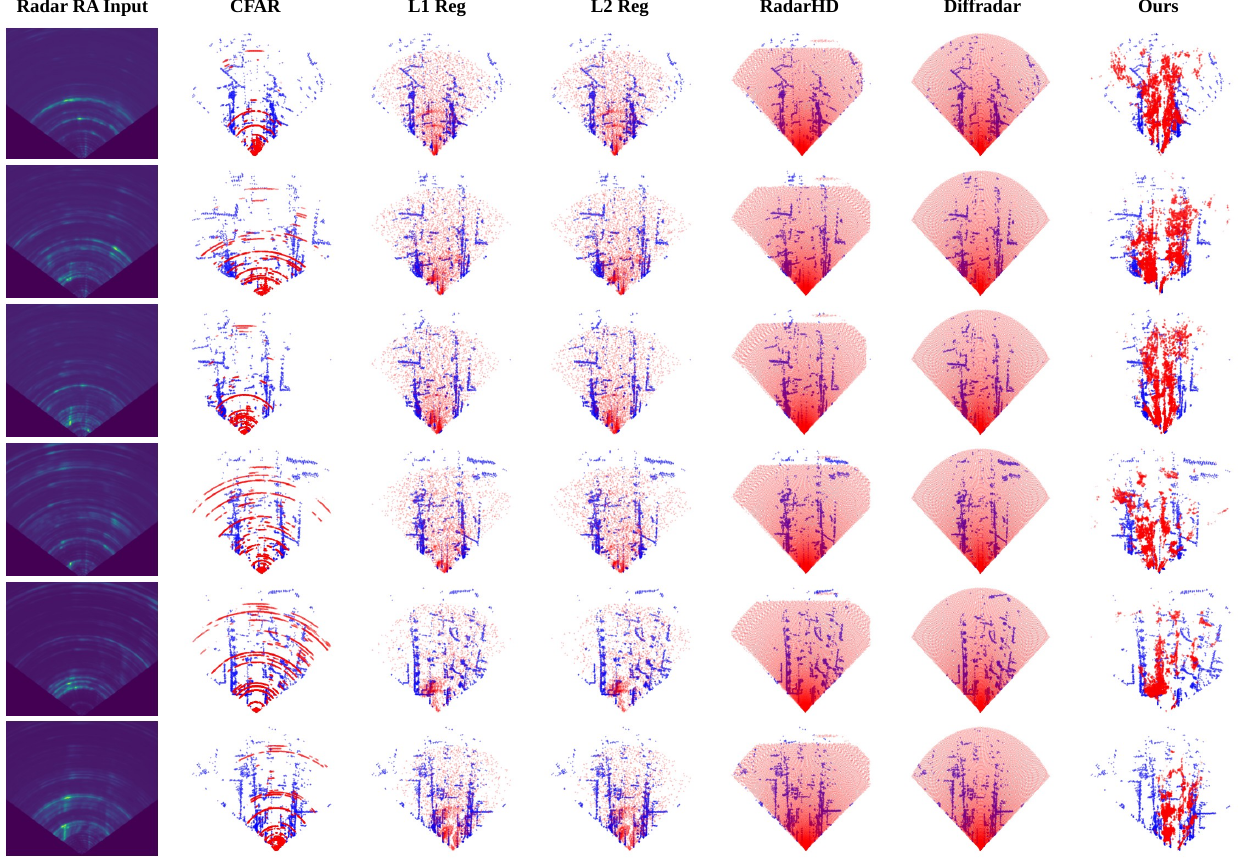}
    \caption{Cross-dataset comparison of radar point cloud enhancement of different methods on K-Radar dataset.}
    \label{fig:appendix_vis_kradar}
\end{figure}

\begin{figure}[htb]
    \centering
    \includegraphics[width=1.0\linewidth]{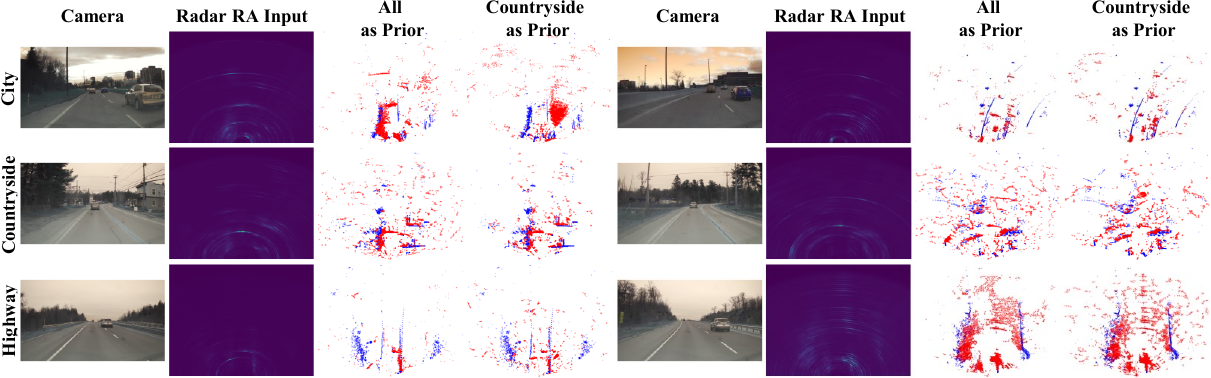}
    \caption{Cross-scenario comparison of radar point cloud enhancement of different methods on RADIal dataset.}
    \label{fig:appendix_vis_cross_scenario}
\end{figure}

We depict more visualization results in this section including point cloud enhancement visualization results of RADIal dataset in Figure \ref{fig:appendix_vis_radial}, cross-dataset visualization results of K-Radar dataset in Figure \ref{fig:appendix_vis_kradar}, and cross-scenario visualization results of RADIal dataset in Figure \ref{fig:appendix_vis_cross_scenario}.

\newpage
\clearpage

\end{document}